\documentclass[runningheads]{llncs}
\usepackage{amsmath}
\usepackage[T1]{fontenc}
\usepackage{graphicx}
\usepackage{tabularx,booktabs}
\newcolumntype{C}{>{\centering\arraybackslash}X} 
\begin{document}
\title{Realistic Surgical Image Dataset Generation Based On 3D Gaussian Splatting}
\titlerunning{Surgical Dataset Generation Based on 3DGS}
%
\author{Tianle Zeng\inst{1}\orcidID{0009-0004-1659-6643} \and
Gerardo Loza Galindo\inst{1}\orcidID{0000-0003-2841-0506} \and
Junlei Hu\inst{1}\orcidID{0000-0001-7394-5580}\and
Pietro Valdastri\inst{1}\orcidID{0000-0002-2280-5438}\and
Dominic Jones\inst{1}\orcidID{0000-0002-2961-8483}}
\authorrunning{T.Zeng et al.}

%
\institute{University of Leeds, Leeds, UK }
%
\maketitle              
\begin{abstract}
Computer vision technologies markedly enhance the automation capabilities of robotic-assisted minimally invasive surgery (RAMIS) through advanced tool tracking, detection, and localization. However, the limited availability of comprehensive surgical datasets for training represents a significant challenge in this field. This research introduces a novel method that employs 3D Gaussian Splatting to generate synthetic surgical datasets. We propose a method for extracting and combining 3D Gaussian representations of surgical instruments and background operating environments, transforming and combining them to generate high-fidelity synthetic surgical scenarios. We developed a data recording system capable of acquiring images alongside tool and camera poses in a surgical scene. Using this pose data, we synthetically replicate the scene, thereby enabling direct comparisons of the synthetic image quality (27.796$\pm$1.796 PSNR). As a further validation, we compared two YOLOv5 models trained on the synthetic and real data, respectively, and assessed their performance in an unseen real-world test dataset. Comparing the performances, we observe an improvement in neural network performance, with the synthetic-trained model outperforming the real-world trained model by 12\%, testing both on real-world data.

\end{abstract}

\section{Introduction}
Detecting and tracking surgical instruments are crucial data sources used in the automation of Robotic-Assisted Minimally Invasive Surgery (RAMIS), providing essential proprioceptive data to the robot \cite{moccia2020supervised,hasan2021detection,lee2020evaluation}.
The robot's forward kinematics give a general measure of tool pose; however, the inherent compliance in cable-driven surgical robotic mechanisms, designed to ensure surgical safety and adaptability, can introduce positional inaccuracies, complicating the precise tracking and detection of instrument end-effectors \cite{attanasio2021autonomy}.
Computer vision technologies offer a solution to this inaccuracy; however, the lack of high-quality labeled data available in surgical settings often complicates the training and supervision of learning-based methods.
\\
Previous studies \cite{azagra2023endomapper,chlap2021review} have explored creating artificial surgical images from real-world images to combat data scarcity. Using game engines \cite{ozturk2021real,tsirikoglou2020survey} offers a scalable, noise-free solution but fails to accurately replicate real surface properties and textures. Generative neural networks can produce fully synthetic datasets of surgical scenes, including tools, by training on surgical environments \cite{armanious2020medgan,colleoni2021robotic,kazeminia2020gans,ozawa2021synthetic,colleoni2022ssis}. However, this method has limitations: the position and pose of instruments in generated images are fixed and uneditable, it lacks scalability as different scenes require retraining the whole network, and annotating datasets remains a time-consuming and complex task with few methods providing corresponding annotation information.
\\
Neural Radiance Fields (NeRF) \cite{mildenhall2021nerf} construct an implicit 3D model of a scene from photographs taken at known positions, enabling the rendering of 2D images from new, unseen viewpoints, thus generating diverse image datasets.
EndoNeRF \cite{wang2022neural} first applied NeRF in surgery, removing instruments from dynamic videos of soft tissue manipulation to reveal unobstructed tissue images.
Psychogyios et al. \cite{psychogyios2023realistic} trained a light source location-conditioned NeRF to encapsulate a colon sequence's 3D and color information, generating new image datasets. NeRF-based methods surpass generative neural networks in image quality and dataset diversity by producing 2D images from various viewpoints.
However, images generated by these methods do not include surgical instruments, as their removal is a prerequisite for operation. Consequently, such outputs are unsuitable for direct use in neural network training.
\\
The 3D Gaussian Splatting \cite{kerbl20233d} method advances the concept of scene representation and rendering by offering a novel approach to modeling 3D scenes from a collection of images.
This method emphasizes explicit representation and high-quality real-time rendering, allowing for generating detailed and photo-realistic images from new viewpoints \cite{chen2024survey}.
The high-quality image rendering, explicit scene representation, and rapid training times of 3D Gaussian Splatting position it as a potential method to overcome the drawbacks of image dataset generation methods based on NeRF.

\begin{figure}\centering
\includegraphics[width=\textwidth]{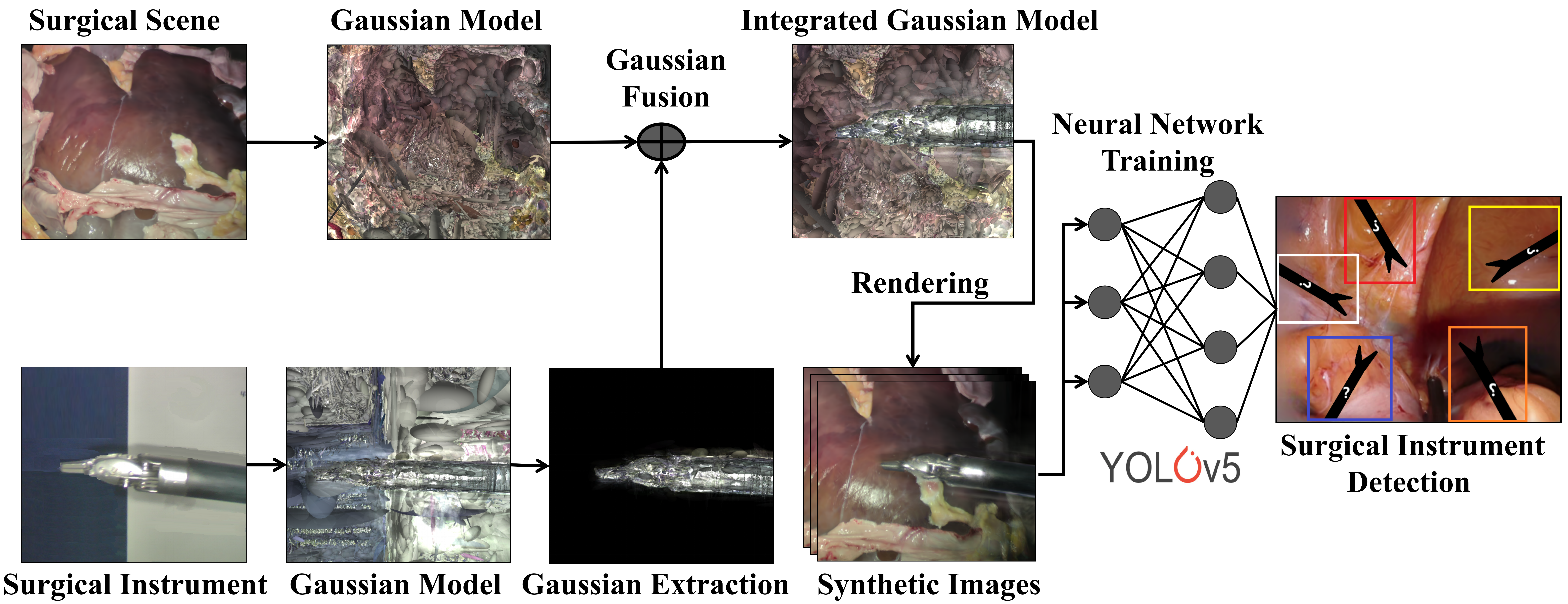}
\caption{Pipeline of proposed method.} \label{fig:pipeline}
\end{figure}

Our approach leverages 3D Gaussian Splatting to produce a novel method that enables the creation of image datasets featuring various surgical instruments across different surgical scenes. The method offers capabilities for scene editing and novel scene synthesis to enhance dataset diversity.
Our key contributions are as follows: 
1) We are the first to apply 3D Gaussian Splatting for medical dataset generation, offering a new methodological route for creating a surgical image dataset. 
2) We developed a technique for precise editing of 3D Gaussian models, allowing independent training and a flexible combination of surgical scenes and instrument models. 
3) The proposed method can automatically generate accurate annotation information alongside image datasets.
4) We demonstrate the high quality of the synthetic image datasets produced by our method and their potential for neural network training.

\section{Methods}
As shown in Figure 1, our method first requires a set of images of the surgical scene and surgical instruments with known camera intrinsic and extrinsic parameters, obtainable through tracking the camera's motion path or using structure from motion (SFM) methods like COLMAP \cite{schonberger2016structure}.
Then, we separately train 3D Gaussian representations for the surgical scene and instruments.
Next, we extract the Gaussian representation of the surgical instruments from the background and perform necessary edits, such as translation and rotation.
Following this, we fuse the instrument's Gaussian representation with that of the surgical scene, resulting in a scene that includes the surgical instruments.
Utilizing the fused Gaussian scene enables the rendering of 2D images with varying poses from multiple viewpoints, facilitating the creation of an image dataset for neural network training.

\subsection{Preliminary: Gaussian Splatting} 
3D Gaussian Splatting \cite{kerbl20233d} is a technique for representing static 3D scenes, distinguished by its differentiability and the ease with which it can be projected into 2D splats. 
This feature enables efficient $\alpha$-blending for rapid image rendering. 
The 3D scenes are represented by a collection of 3D Gaussians defined by a mean $\mathbf{\mu}$ and covariance matrix $\mathbf{\Sigma}$, described in the equation:

\begin{equation}
    \mathrm{G}(x)=\frac{1}{(2 \pi)^{3 / 2}|\Sigma|^{1 / 2}} e^{-\frac{1}{2}(x-\mu)^{\mathrm{T}} \Sigma^{-1} \frac{1}{2}(x-\mu)}
\end{equation}

Where $\Sigma$ is decomposed into rotation matrix $R$ and scaling matrix $S$, writing as $\mathbf{\Sigma} = \mathbf{RSS^{T}R^{T}}$.
The 3D Gaussians are enhanced with opacity and spherical harmonic (SH) coefficients for color representation, enabling the depiction of anisotropic appearances.
These Gaussians encapsulate the 3D spatial information of scenes through learned attributes, which are refined during the training process. 
Gaussian density control step is also implemented to interleave these Gaussians effectively.
Our work is based on 3D Gaussian Splatting, wherein we initially train two distinct Gaussian models: one for the surgical scene and another for the surgical instrument.

\subsection{Gaussian Extraction and Labelling}
After training, we obtain Gaussian models for both the surgical scene and instruments, with the instrument model capturing both the tool and its background.
For Gaussian model fusion, we aim to isolate and use only those Gaussians representing the instruments, necessitating a method to segment these from the background representation.
In our instrument Gaussian model, the tool Gaussians are densely centered with sparse background Gaussians distributed in the distance. This allows for extraction by selecting center-distributed Gaussians and filtering out others.

\begin{figure}\centering
\includegraphics[width=\textwidth]{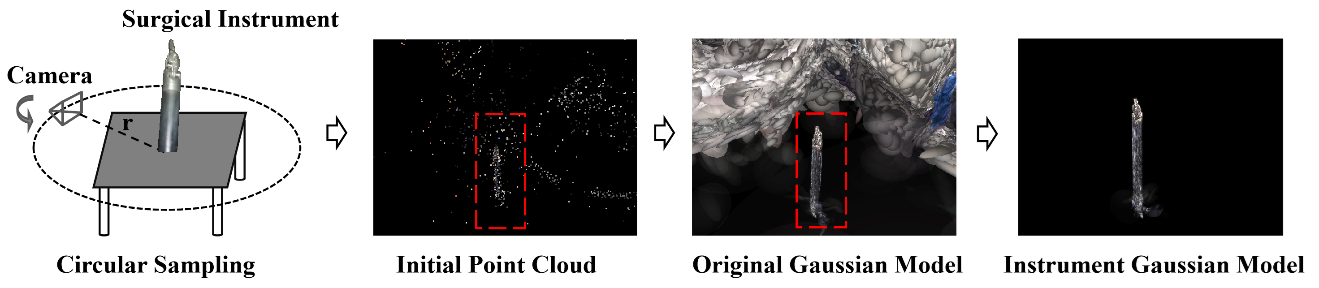}
\caption{Using circular sampling, we center the surgical tool in the scene, resulting in a dense distribution at the center (highlighted by a red rectangle).} \label{fig:tool_scan}
\end{figure}

Labeling Gaussian models representing surgical instruments is crucial for the subsequent fusion process, where adjustments to the Gaussians' orientations and positions are required. Although conceptually simple, this labeling necessitates significant changes to the program's data structures, making it complex and labor-intensive.
An efficient alternative uses existing Gaussian properties, notably the color difference between instruments and scenes, as natural labels. In Gaussians, color is represented using the Spherical Harmonics (SH) function \cite{kerbl20233d}. Therefore, instruments can be labeled by tagging the SH function's direct current component within the Gaussians, generating a secondary tool representation solely for data labeling.

\subsection{Gaussian Scene Fusion}
Once we have obtained 3D Gaussian representations of the surgical scene and the segmented surgical instruments, we can then combine the two to produce a fully synthetic scenario with a specified tool pose.
The representation of 3D Gaussians is independent; as long as the properties of an individual Gaussian are not altered, adding new Gaussians to a Gaussian model does not affect its representation \cite{kerbl20233d}.
Therefore, we can achieve Gaussian scene fusion by incorporating the extracted surgical instrument into the surgical scene.
However, the fused model at this stage is not yet suitable for generating synthetic images.
For each synthetic image, we must transform the instrument Gaussian to match its intended location with respect to the camera and background.
3D Gaussians are represented by their mean $\mathbf{\mu}$ and covariance $\mathbf{\Sigma}$, indicating their position and orientation \cite{kerbl20233d}, which can be adjusted to correct the Gaussians' pose in the fused scene as: 

\begin{equation}
\begin{split}
    &G^{\prime}(x)=\frac{1}{(2 \pi)^{3 / 2}|\Sigma \prime|^{1 / 2}} e^{-\frac{1}{2}\left(x-\mu^{\prime}\right)^T \Sigma^{\prime-1}\left(x-\mu^{\prime}\right)} \\
    &where:  \mu' = \mu + \Delta\mu; \Sigma' = R\Sigma R^T
\end{split}
 \end{equation}

Where $\Delta\mu$ denotes the translation of Gaussians, $R$ denotes the rotation of Gaussians.
By editing the fused Gaussians, we can generate synthetic images in arbitrary orientations and positions of the camera and instrument.
\subsection{Automatic Annotation Generation}
As an extension of the fused Gaussian representation, we can generate a pixel-wise segmentation mask that clearly delineates tool boundaries in synthetic images. By exclusively rendering the tool Gaussian, labeled accordingly, we utilize the differential Gaussian rasterization pipeline introduced by \cite{kerbl20233d}. These 3D Gaussians are projected into 2D using the covariance matrix $\Sigma'$:
\begin{equation}
\Sigma' = J W \Sigma W^T J^T,
\end{equation}
where $J$ is the Jacobian from the affine approximation of the projective transformation, $W$ is the view matrix for world-to-camera coordinates, and $\Sigma$ is the 3D covariance matrix. Pixel colors on the image plane, denoted by $C$, are computed by \(\alpha\)-blending the contributions of Gaussians ordered from nearest to farthest:
\begin{equation}
C = \sum_{i \in N}\alpha_i c_i\prod_{j=1}^{i-1} (1 - \alpha_j),
\end{equation}
\begin{equation}
\alpha_i = \sigma_i e^{-\frac{1}{2} (\mu - u_i)^T \Sigma' (\mu - u_i)},
\end{equation}
where $c_i$ is the color of each Gaussian, and $u_i$ represents their projected $uv$ coordinates. Rendering only the tool Gaussians, the background appears black ($C=0$). In the resulting 2D image, only the surgical instrument regions are colored, enabling clear segmentation. By setting a contrast threshold, we differentiate the black background (background) from the colored instrument (foreground) to generate the 2D mask. This mask allows for the application of contour detection algorithms to define the contours of the foreground. The pixel coordinates of these bounding boxes facilitate the automated creation of annotation files.


\subsection{Experimental Data Recording}
\begin{figure}\centering
\includegraphics[width=.9\textwidth]{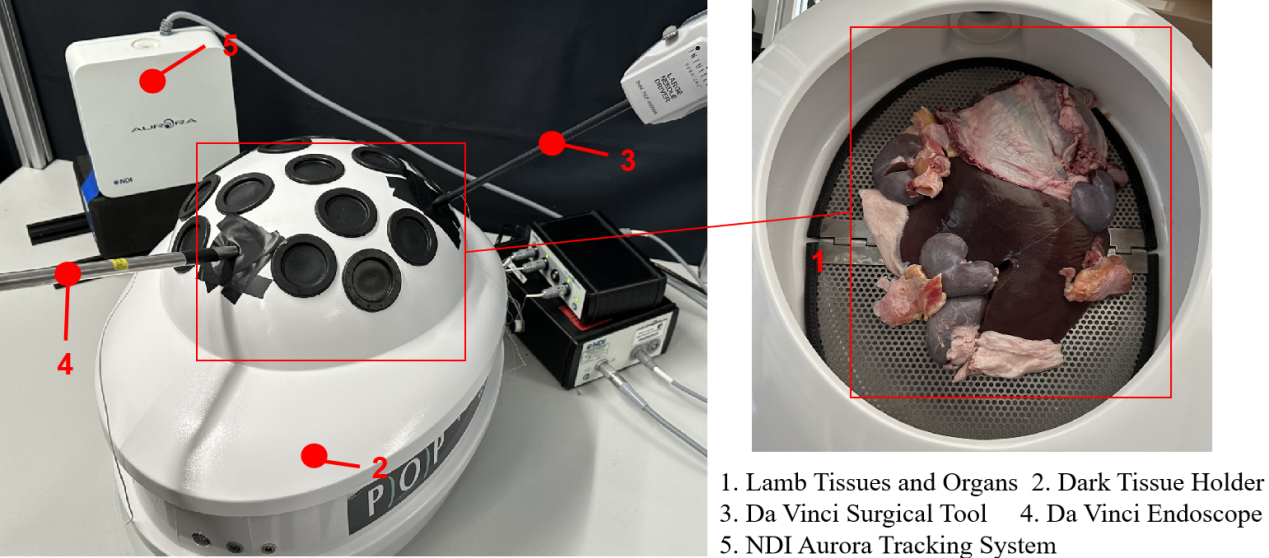}
\caption{Our dataset recording platform.} \label{fig:data_rec}
\end{figure}

In order to assess our synthetic images, we present a novel methodology for acquiring images and instrument poses (camera and tools) from real scenarios. Images with tools are used for validation, but realistic backgrounds and tools are not mixed during the generation of Gaussian representation.
Fresh ex-vivo lamb liver, kidney, and fat placed within a laparoscopic trainer platform (POP Trainer, Optimist GMBH, Austria) to mimic the surgical scene.
We use a clinical laparoscopic camera from the daVinci Surgical System (HD-2 Stereoendoscope Module, Intuitive Surgical) for image collection, and a daVinci Large Needle Driver as our tool.
Given the current requirement for a rigid tool, we fixed all joints in a neutral position. 
We utilise the NDI Aurora electromagnetic tracking system with 6DoF sensors for our Ground Truth pose of the tool and camera. 
Image and pose data were collected synchronously.
The setup for surgical data acquisition is depicted in Figure \ref{fig:data_rec}.

In the dataset, we recorded three distinct videos. 1) A training dataset for representing the background scene with no tools present; 2) An isolated acquisition of the surgical tools; and 3) A ground truth dataset containing both background and tool, designated below as the Ground Truth (GT) dataset.
An additional test dataset mimicking the GT dataset was also recorded.

\section{Experiments and Results}
After acquiring all three data sets (Background, Tool, GT), we utilized the tool and camera positions within the GT dataset as direct inputs to the image synthesis.
This approach guarantees that generating images that align with the GT images regarding the same camera and surgical instrument positions is feasible, replicating the GT image with a synthesized version.
Consequently, real-world images can be a test case for synthetic images under such conditions.

\subsection{Synthetic Image Quality Evaluation}
\begin{figure}\centering
\includegraphics[width=\textwidth]{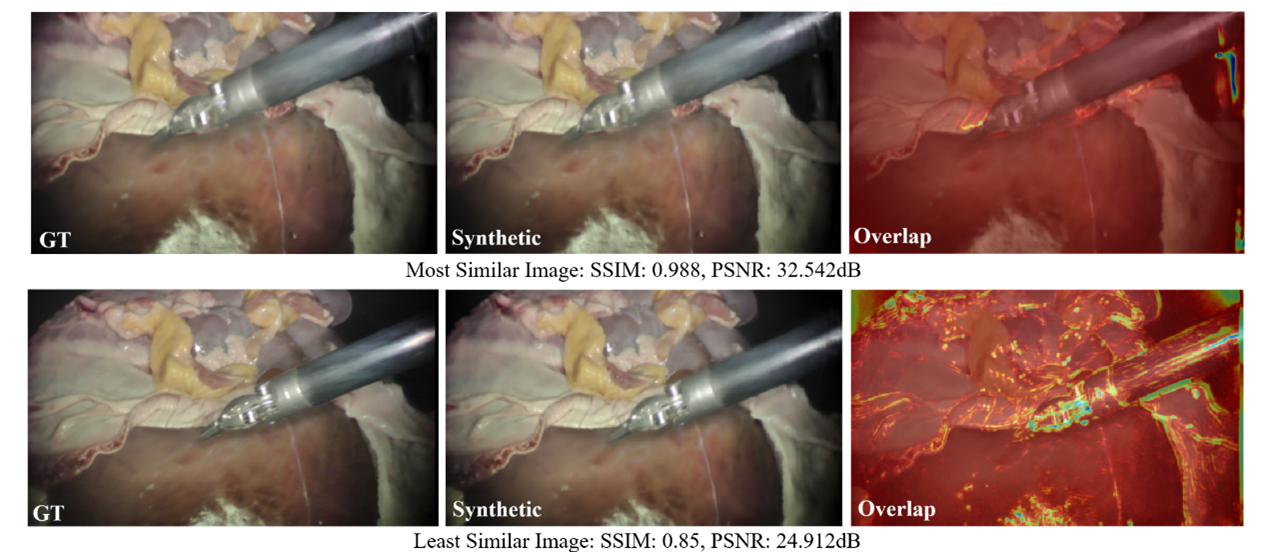}
\caption{Illutsration of GT and Synthetic pairs for the most and least similar image, indicating the regions of increased difference} \label{fig:comparison}
\end{figure}

A key challenge in image generation is obtaining precise GT images for direct comparison. We utilize Gaussian editing and tool pose tracking to acquire accurate GT images, a capability that sets our method apart from other generative approaches, which often lack precise GT data. For evaluation, we focus on comparing our results with these GT images to accurately assess our method's effectiveness. We evaluate the quality of our synthesized images using Peak Signal-to-Noise Ratio (PSNR), Structural Similarity Index Measure (SSIM), and Learned Perceptual Image Patch Similarity (LPIPS). We also overlay synthetic and GT images to identify discrepancies. The results, illustrated in Figure \ref{fig:comparison}, demonstrate the superiority of our method in producing high-quality images in GT scenes compared to alternatives. We conducted rigorous comparisons with two state-of-the-art (SOTA) Nerf-based methods \cite{mildenhall2021nerf,muller2022instant,tancik2023nerfstudio}, chosen for their comparable training durations. The PSNR, SSIM, and LPIPS scores are presented in Table \ref{tab1}.

\begin{table}
\centering
\caption{Comparative Analysis of Image Quality of GT Scene (Mean and standard deviation).}\label{tab1}
\begin{tabularx}{0.8\textwidth}{@{} l *{3}{C} c @{}}
\toprule
Method & PSNR $\uparrow$ & SSIM $\uparrow$ & LPIPS $\downarrow$\\
\midrule
Instant-NGP & 16.603$\pm$0.782 & 0.741$\pm$0.015 & 0.758±0.047\\
Nerfacto & 22.736$\pm$1.435 & 0.796$\pm$0.016 & 0.394$\pm$0.035\\
Ours & 27.796$\pm$1.796 & 0.912$\pm$0.029 & 0.287$\pm$0.022\\
\bottomrule
\end{tabularx}
\end{table}

\subsection{Neural Network Training Experiment}

Next, we evaluated the efficacy of our synthesized images for training neural networks, specifically using YOLOv5 \cite{redmon2016you} for object detection. We trained two models: one on the GT Dataset and one on our synthetic data, both using 1568 images. For consistency, images were resized to 640x640 pixels, and training was conducted for 100 epochs using YOLOv5's default parameters. All training was performed on an NVIDIA GeForce RTX 4050 (6G). Our test set comprised 300 images, each with unique combinations of camera and instrument poses. We assessed precision and recall to determine the viability of training with synthesized data and conducted multi-fold experiments for reliability. Table \ref{tab2} shows our results, indicating that models trained on generated images outperform those trained on GT images in both precision and recall. This improvement is attributed to our method's ability to augment data by rendering images from new viewpoints and instrument poses, providing stronger priors for detection.

\begin{table}
\centering
\caption{Performance Comparison of Neural Networks Trained with Synthetic vs. Ground Truth Images on the 300 image real-world Test Dataset}\label{tab2}
\begin{tabularx}{0.6\textwidth}{@{} l *{2}{C} c @{}}
\toprule
Model & Precision $\uparrow$ & Recall $\uparrow$ \\
\midrule
Synthetic Training Input & 0.801 & 0.901\\
GT Training Input & 0.703 & 0.804\\
\bottomrule
\end{tabularx}
\end{table}


\section{Conclusion}

This paper introduces a novel surgical image dataset generation method based on 3D Gaussian Splatting, aiming to address the challenge of insufficient surgical image datasets.
We first trained Gaussian models representing surgical scenes and instruments separately to achieve this.
We adopted a circular sampling strategy for the surgical scene Gaussian models, enabling accurate extraction and labeling of surgical instrument Gaussians.
We created new scene models by fusing the extracted surgical instrument Gaussians with those from the surgical scene, allowing for image rendering of surgical instruments in any pose.
This process also auto-generates annotation information for surgical instruments.
Our experiments confirmed the high quality of images generated by our method, achieving a PSNR of 29.592.
Our generated datasets have been proven effective for training neural networks, resulting in a 12\% improvement in performance when models are trained on generated images compared to those trained on ground truth images.
Currently, our method can generate and edit static image data of surgical tools within static scenes.
This work hopes to alleviate the data scarcity in the surgical domain and inspire further enhancement of 3D Gaussian Splatting techniques for data generation.

\begin{credits}

\subsubsection{\discintname}
The authors have no relevant financial or non-financial interests to disclose.
\end{credits}

\bibliographystyle{unsrt}
\bibliography{refs}
\end{document}